\documentclass[conference]{IEEEtran}
\IEEEoverridecommandlockouts
\usepackage{cite}
\usepackage{amsmath,amssymb,amsfonts}
\usepackage{algorithmic}
\usepackage{balance}
\usepackage{graphicx}
\usepackage{textcomp}
\usepackage{tabularx}
\usepackage{xcolor}
\usepackage{multirow}
\usepackage{subfigure}
\def\BibTeX{{\rm B\kern-.05em{\sc i\kern-.025em b}\kern-.08em
    T\kern-.1667em\lower.7ex\hbox{E}\kern-.125emX}}

\title{\textsc{GiNet}: Integrating Sequential and Context-Aware Learning for Battery Capacity Prediction\\
\thanks{This work was supported in part by SIT’s Ignition Grant (STEM) (Grant ID: IG (S) 2/2023 – 792), A*STAR under its MTC Programmatic (Award M23L9b0052), MTC Individual Research Grants (IRG) (Award M23M6c0113), and the National Research Foundation, Singapore and Infocomm Media Development Authority under its Future Communications Research \& Development Programme (Grant FCP-SIT-TG-2022-007).}
}

\makeatletter
\newcommand{\linebreakand}{%
  \end{@IEEEauthorhalign}
  \hfill\mbox{}\par
  \mbox{}\hfill\begin{@IEEEauthorhalign}
}
\makeatother

\author{\IEEEauthorblockN{Sara Sameer}
\IEEEauthorblockA{
\textit{Singapore Institute of Technology}
\\\textit{Singapore, 828608}
\\sara.sameer@singaporetech.edu.sg
}
\and
\IEEEauthorblockN{Wei Zhang$^*$\thanks{$^*$ Corresponding author. Email: wei.zhang@singaporetech.edu.sg.}}
\IEEEauthorblockA{
\textit{Singapore Institute of Technology}
\\\textit{Singapore, 828608}
\\wei.zhang@singaporetech.edu.sg
}
\and
\IEEEauthorblockN{Xin Lou}
\IEEEauthorblockA{
\textit{Singapore Institute of Technology}
\\\textit{Singapore, 828608}
\\xin.lou@singaporetech.edu.sg
}
\linebreakand
\IEEEauthorblockN{Qingyu Yan}
\IEEEauthorblockA{
\textit{Nanyang Technological University}
\\\textit{Singapore, 639798}
\\alexyan@ntu.edu.sg
}
\and
\IEEEauthorblockN{Terence Goh}
\IEEEauthorblockA{
\textit{ST Engineering}
\\\textit{Singapore, 619523}
\\kenglengterence.goh@stengg.com
}
\and
\IEEEauthorblockN{Yulin Gao}
\IEEEauthorblockA{
\textit{ST Engineering}
\\\textit{Singapore, 619523}
\\gao.yulin@stengg.com
}
}

\begin{document}
\bstctlcite{IEEEexample:BSTcontrol}

\maketitle

\begin{abstract}
The surging demand for batteries requires advanced battery management systems, where battery capacity modelling is a key functionality. In this paper, we aim to achieve accurate battery capacity prediction by learning from historical measurements of battery dynamics. We propose \textsc{GiNet}, a \underline{g}ated recurrent units enhanced \underline{I}nformer \underline{net}work, for predicting battery's capacity. The novelty and competitiveness of \textsc{GiNet} lies in its capability of capturing sequential and contextual information from raw battery data and reflecting the battery's complex behaviors with both temporal dynamics and long-term dependencies. We conducted an experimental study based on a publicly available dataset to showcase \textsc{GiNet}'s strength of gaining a holistic understanding of battery behavior and predicting battery capacity accurately. \textsc{GiNet} achieves 0.11 mean absolute error for predicting the battery capacity in a sequence of future time slots without knowing the historical battery capacity. It also outperforms the latest algorithms significantly with 27\% error reduction on average compared to Informer. The promising results highlight the importance of customized and optimized integration of algorithm and battery knowledge and shed light on other industry applications as well.
\end{abstract}

\begin{IEEEkeywords}
battery capacity, state of charge, machine learning, industrial artificial intelligence
\end{IEEEkeywords}

\section{Introduction}
Batteries power devices and systems for different industry sectors, from consumer electronics to mobility. They are often considered the cornerstone of modern technology and society. The demand for Li-ion batteries alone is expected to grow by over 30\% annually to reach $\sim$4.7 TWh by 2030. Batteries for mobility, e.g., electric vehicles (EVs), account for about 90\% of the demand \cite{fleischmann2023battery}. In line with the surging demand, the revenue of the battery value chain is expected to surpass 400 billion dollars by 2030. The huge demand and broad adoption require batteries to be safe, efficient, and sustainable, typically ensured by the critical battery management system (BMS). A BMS shall actively monitor battery dynamics, e.g., current, voltage, and temperature, and guard against overcharging, deep discharge, thermal runaway, etc., that can result in hazards like fires and explosions \cite{10710765}. Often, BMS makes decisions based on several important indicators. One of them is state of charge (SoC), which tells the remaining capacity of a battery as the ratio of its total capacity at any given time \cite{10258162}. SoC is highly important for battery management, especially in applications like EVs and smart grids. 

Accurate measurement of SoC has been a challenging task. Traditional methods include open-circuit voltage \cite{CHEN2019113758}, Coulomb-counting \cite{MOHAMMADI2022104061}, etc. These methods are simple and widely used, though they face challenges with limited accuracy and the inability to adapt to dynamic battery conditions. For example, \cite{CHEN2019113758} requires batteries to be at rest for accurate measurements and is impractical for applications during operation time. Data-driven approaches based on machine learning (ML) have gained more significance in recent years due to their high accuracy and robustness. Battery measurements are often sequential and naturally time-series ML algorithms, e.g., recurrent neural networks (RNNs), are used for SoC estimation and prediction. Long short-term memory (LSTM) is perhaps the most popular type of RNN and serves as a baseline in many SoC studies. Newer RNNs have been investigated recently to improve SoC prediction performance and a comparison study is available in \cite{VIDAL2022103660}. The latest Transformer-based ML is also capable of handling sequential data but in a different way than RNNs, e.g., with attention mechanisms. The Transformer models have demonstrated superior performance to traditional RNNs for many applications \cite{LIN2022111}. In a recent study \cite{10382587}, a temporal transformer-based network is used to model the relationship between battery measurements and SoC. A common feature of most ML-based SoC studies is that each uses a single ML algorithm. Battery behavior is known to be complex, diverse, and multi-faceted, and single-model-based SoC prediction is often insufficient to understand battery behavior well and deliver optimal performance. 

SoC models based on multiple ML algorithms are promising to capture the battery behavior complexity by leveraging different algorithms' complementary strengths. In \cite{8754752}, LSTM is integrated with a convolutional neural network (CNN) to estimate SoC. However, the idea is to process battery data of different modalities instead of focusing on the application and exploiting complex battery dynamics. Looking into the battery capacity, there are different aspects of the battery behaviors and operating conditions. Naturally, there is sequential information in the battery data, in time series. The information is related to the battery's temporal dynamics, e.g., gradual capacity decline during short-term discharging within a cycle and long-term battery degradation across cycles. There also exists contextual information and battery capacity related external factors, such as ambient temperature, fast charging, high discharge rates, and frequent deep discharge. These factors affect battery capacity in ways that are not fully time-dependent, or sequential. Capturing such contextual information complements sequential learning to gain a holistic understanding of battery behavior and accordingly, predict SoC accurately.

In this paper, we aim to integrate sequential and context-aware learning for battery capacity prediction. RNNs are natural options to process sequential information. Specifically, we employ gated recurrent units (GRU), a variant of LSTM with a simplified architecture and enhanced generalizability \cite{chung2014empiricalevaluationgatedrecurrent}. Contextual information is not always temporal and we apply Transformer models to capture long-range dependencies with their powerful attention mechanisms. Informer \cite{Zhou_Zhang_Peng_Zhang_Li_Xiong_Zhang_2021}, one of the latest Transformer architectures, is used in this study. Informer has efficient attention and sequence distillation mechanisms that make it suitable for processing long-sequence data and extracting contextual information effectively. Accordingly, we propose \textsc{GiNet}, a \underline{G}RU-enhance \underline{I}nformer \underline{net}work for battery capacity prediction. \textsc{GiNet} configures Informer to be the core architecture for battery analytics. We customize and optimize Informer's embedding layer with GRU-generated features and feature fusion to provide enhanced embedding with both sequential and contextual information. The information is analyzed by Informer's encoder and decoder to generate battery SoC prediction. In summary, we have the following main contributions in this paper.
\begin{itemize}
    \item We designed and developed \textsc{GiNet}, a GRU-enhanced Informer architecture for battery capacity prediction based on historical battery monitoring data.
    \item We customized and optimized the integration GRU and Informer to capture both sequential and contextual features of batteries effectively with feature fusion.
    \item We conducted an experimental study and showed that \textsc{GiNet} predicts battery SoC accurately with a minimal mean absolute error (MAE) of 0.11 and achieves 76\% and 27\% performance improvement compared to GRU and Informer, respectively.
\end{itemize}

The rest of the paper is organized as follows. We present \textsc{GiNet} for battery capacity prediction in Section \ref{sec:sys}. In Section \ref{sec:exp}, we conduct an experimental study and present results and discussions. Finally, we conclude this paper and suggest future works in Section \ref{sec:conclusion}.

\section{\textsc{GiNet} Methodology}
\label{sec:sys}
\begin{figure}
    \centering	\includegraphics[width=0.99\linewidth]{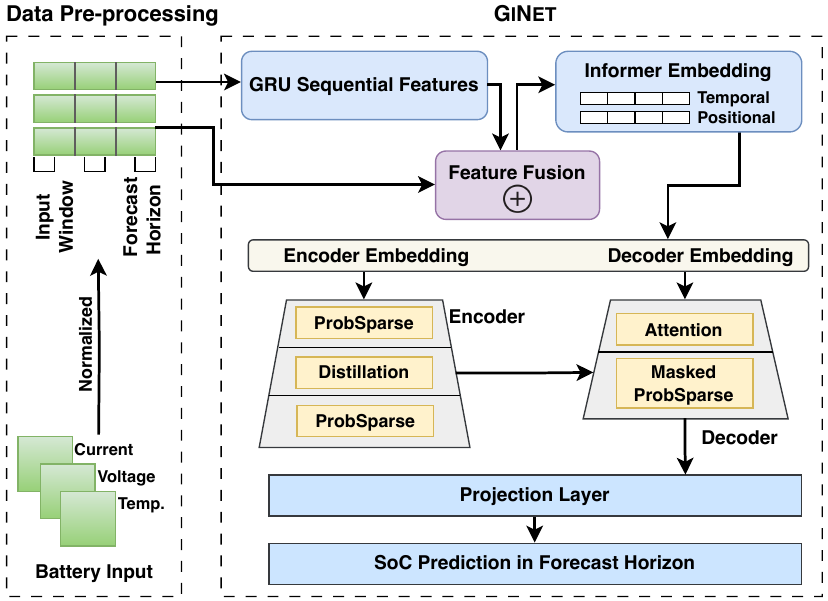}
    \caption{An illustration of \textsc{GiNet} architecture. \textsc{GiNet} performs data pre-processing for the battery time-series data. GRU is used to capture sequential information and extract corresponding features, and Informer is employed to model long-term dependence and extract contextual information. Supporting modules include feature fusion, output mapping to forecast horizon, etc.}
    \label{fig:ginet}
\end{figure}

We present the overall architecture and main components of \textsc{GiNet} in this section. An illustration of the \textsc{GiNet} architecture is in Fig. \ref{fig:ginet}.

\subsection{Model Overview and Technology Recap}
\textsc{GiNet} uses the battery monitoring data for battery capacity prediction. The collected raw data undergoes data pre-processing to generate input vectors ready for \textsc{GiNet} prediction. The input of \textsc{GiNet} is first processed by a GRU module to extract sequential features and understand time-series patterns from the battery data. The GRU features are fused with the original feature to enhance the sequential features with contextual information on battery dynamics. The fused features serve as the input of the Informer's embedding and Informer analyzes the features with its attention-based encoder and decoder. Finally, \textsc{GiNet} maps the output of the Informer module to the battery capacity forecast horizon and produces the SoC prediction results accordingly.

Before presenting \textsc{GiNet}'s technology details, we provide a brief recap of its supporting ML technologies, GRU and Informer, as below.

\subsubsection{GRU}
GRU is one type of RNN. Compared to the popular RNN-based LSTM, GRU's structure is simple. GRU has one less gate than LSTM. It combines LSTM's forget and input gates into an update gate to control the amount of historical information to be retained and new information to be used. It also has a reset gate to control the historical information to be deleted and GRU's unique gate mechanism minimizes gradient vanishing issues that are common for other RNNs. Finally, the gate output is used to calculate the final hidden state, which is the GRU-based representation of the input data, e.g., GRU-based battery temporal dependencies. GRU details such as gates and hidden state calculations are available in \cite{chung2014empiricalevaluationgatedrecurrent}.

\subsubsection{Informer}
Transformers have been widely used for time-series forecasting applications and many of them are based on full attention as,
\begin{equation}
    \texttt{Attention}(Q, K, V) = \texttt{softmax}\left(QK^T / \sqrt{d}\right)V
    \label{eq:attention}
\end{equation}
where $Q$, $K$, and $V$ are query, key, and value matrices, respectively, and $d$ is the input dimension. A key challenge of Eq. (\ref{eq:attention}) is that the attention weights of all pairs of queries and keys are computed with a high time complexity of $O(n^2)$ with an input length of $n$. Informer aims to process long sequence data effectively. Its key innovation is to reduce computation complexity with a new attention mechanism called \texttt{ProbSparse} self-attention as, 
\begin{equation}
    \texttt{ProbSparse}(Q, K, V) = \texttt{softmax}\left(\overline{Q}K^T / \sqrt{d} \right)V
    \label{eq:probsparse}
\end{equation}
where $\overline{Q}$ is a sparse matrix which contains top-$u$ queries with dominant attention only and $u$ is expected to be much smaller than $n$. The dominance of each query key pair is quantified by a query sparsity measurement function $M(q, K)$ for a query $q$. The new attention allows the Informer to achieve a time complexity of $O(n \log n)$. Furthermore, Informer uses distillation operations to produce shortened representations of long sequences. Overall, Informer with its attention-based encoder and decoder processes data input and performs linear transformations for forecast generation. Informer details such as sparsity measurements can be found in \cite{Zhou_Zhang_Peng_Zhang_Li_Xiong_Zhang_2021}.

\subsection{Building Blocks}
Based on the model overview, we decompose \textsc{GiNet} architecture into several building blocks and present the technical details below.

\subsubsection{Data Pre-processing}
Most ML algorithms favor big data for both quantity and diversity to achieve optimal performance. However, big data may not be practical for real-world applications like battery capacity prediction, as it requires increased costs for data collection, storage, processing, etc. In this paper, we aim to develop a practical solution for batteries and we consider common battery measurements including voltage, current, and temperature that are easy to monitor for \textsc{GiNet} processing. Nevertheless, we would like to mention that \textsc{GiNet} is not restricted to specific features, and its input can be customized to accommodate different battery systems.

Specifically, let $\mathbf{x}_t = (x_t^\mathcal{I}, x_t^\mathcal{V}, x_t^\mathcal{T})$ be the battery data at time slot $t$ where $\mathcal{I}$, $\mathcal{V}$, and $\mathcal{T}$ represent current, voltage, and temperature, respectively. \textsc{GiNet} is capable of processing the data of a sequence of time slots, i.e., time-series, and we apply over-lapping windows \cite{10382587} to prepare the input data of \textsc{GiNet}. Let $T^\text{in}$ be the length of the input window and the input of \textsc{GiNet} is $X_t = (\mathbf{x}_{t-T^\text{in}}, \mathbf{x}_{t-T^\text{in}+1},\ldots,\mathbf{x}_{t-1})$ at time slot $t$. \textsc{GiNet} can estimate the battery capacity of the current time slot as well as future time slots, referred to as forecast horizon. Let $T^\text{out}$ be the length of the forecast horizon and $\hat{y}_t$ be \textsc{GiNet}'s estimated SoC for time $t$. With the input of $T^\text{in}$ time slots data, \textsc{GiNet} produces $\hat{Y}_t = (\hat{y}_t, \hat{y}_{t+1},\ldots,\hat{y}_{t+T^\text{out}-1})$ for $T^\text{out}$ time slots at time $t$. For simplicity, let $\hat{y}_t^\text{SoC} = \sum_{i=t}^{t+T^\text{out}-1}\hat{y}_i / T^\text{out}$ be the average of the $T^\text{out}$ estimations and our reported results in this paper are based on $\hat{y}_t^\text{SoC}$ by default. Battery capacity datasets often provide the ground-truth SoC as $Y_t = (y_t, y_{t+1},\ldots,y_{t+T^\text{out}-1})$ for evaluating the performance of battery capacity models. However, the ground truth cannot be used during model training and is unavailable in many practical settings. Furthermore, we employ min-max normalization for input feature scaling and keep the output non-normalized to align with real-world SoC prediction. 

\subsubsection{GRU-enhanced Features and Feature Fusion}
At each time slot, we extract features from the \textsc{GiNet} input and expect the features to be better correlated with battery capacity than the raw input. We aim to capture both temporal dependencies and contextual features. In \textsc{GiNet}, we employ GRU to extract temporal dependencies from the sequential battery data input. Specifically, GRU processes the sequential input over a sequence of time slots and generates high-dimensional (1024 in this paper) hidden state representation at each time slot. We configure two layers for hierarchical feature extraction and representation where the output of the first layer serves as the input of the second layer and a dropout layer is applied after the first layer to prevent overfitting. The representations effectively summarize the temporal dependencies from the battery time series and encode complex interactions between battery dynamics over time. Finally, the representations are passed through a linear transformation to map the representations to feature space that matches the original input and let GRU's generated feature be $H_t = \texttt{GRU}(X_t)$. Such mapping ensures compatibility for the following feature fusion. 

\subsubsection{Feature Fusion}
We apply an element-wise addition between the GRU-generated features and the \textsc{GiNet}'s original input, and the fused feature is $F_t = H_t + X_t$ for time $t$. We expect the fused features to retain contextual features from the original input while embedding temporal features extracted by GRU, and serve as the input of the Informer-based data modeling detailed below.

\subsubsection{Informer-based SoC Prediction} 
The fused feature $F_t$ is passed to the Informer's embedding layer first. $F_t$ is converted into a high-dimensional space suitable for Informer's attention mechanism. We use three different embedding techniques, including positional, value, and temporal, to enrich the time-series battery data. The first one helps capture the position of each data point of the input and value embedding uses a 1-dimensional convolutional layer to expand the features into the desired embedding dimension. The last one encodes time-related features, e.g., seconds, to capture the potential periodic nature of batteries. Finally, the input from all three embedding techniques is unified into a single data embedding layer.

The following stage of Informer processing is encoding. The key component of encoding is the multi-head self-attention which incorporates \texttt{ProbSparse} attention for processing long sequence data efficiently. The process efficiency can be further improved with a distillation mechanism that reduces sequence length with convolutional layers.  Another component is the feed-forward layer, which expands \textsc{GiNet}'s capacity to learn complex transformations by introducing non-linearity through activation functions like \texttt{ReLU}, further refining the feature space. Overall, the encoder-generated features are expected to be rich in temporal and contextual information, ready to be processed by the decoder for battery analytics.

In the decoding stage, a unique design is to combine the decoder's input and the encoder's output. The decoder first utilizes its own data along with a placeholder sequence representing forecast horizon to perform multi-head \texttt{ProbSparse} self-attention operation. Then, it integrates the encoder's intermediate output where the decoder aligns its predictions with the feature representations extracted by the encoder. This step ensures that the decoder leverages both historical dependencies as well as the temporal and contextual information learned during encoding. Finally, the output of the decoder is passed to a fully connected layer to adjust the dimension of the output to the prediction horizon of battery capacity as $\hat{Y}_t = \texttt{Informer}(F_t)$ at time $t$.

Overall, \textsc{GiNet} infuses GRU into Informer, where GRU captures fine-grained temporal patterns to complement Informer's attention-based long sequence modeling. Such synergy is essential for accurate battery capacity prediction with battery dynamics being both temporal and contextual.

\section{Experimental Study}
\label{sec:exp}
We present our experimental study in this section for experiment setup, results presentation, and discussions. 

\subsection{Data}
A battery capacity model shall be trained and optimized with a real-world battery dataset. For our experimental study, we use an open dataset \cite{kollmeyer2018panasonic}. The dataset is based on Panasonic 18650PF Li-ion batteries and tests were conducted across a wide temperature range, i.e., $-$20 to 25 °C, to analyze battery performance variations under different ambient conditions. The key tests involved in this dataset include charging and discharging batteries at a slow and steady rate (C/20 charge-discharge cycles), measuring battery power and energy capabilities (hybrid pulse power characterization), and analyzing the internal resistances of the cells (electrochemical impedance spectroscopy). The tests also simulated the battery performance of a Ford F150 electric truck under different driving conditions, e.g., aggressive urban driving (US06), highway driving (HWFET), typical urban driving (UDDS), and tailored driving profiles generated by a neural network. The dataset provides high-resolution (10 measurements per second) battery dynamics, including voltage, current, power, amp-hours, watt-hours, and battery temperature. Overall, this data offers a ground for our experimental study with diverse operational conditions for model development and validation.

With the dataset, we parsed, concatenated, normalized, and split the data into training, validation, and test datasets, with a ratio of 10:2:5. The training data is composed of mixed cycles and driving profiles at four different ambient temperatures (i.e., $-$10, 0, 10, and 25 °C). Two cycles are reserved for testing to evaluate model performance with unseen data.

\subsection{Model Configuration and Implementation}
\textsc{GiNet} involves several parameters. We have optimized the parameters and chosen one of the optimal configurations as the default setting. The batch size is 32, and the learning rate is $1\mathrm{e}{-4}$. A scheduler for learning rate decrease, zero-padding, and distillation are set to \texttt{true}. We have 20 epochs and early stopping can be triggered when the model stops improving. Battery capacity modelling is relatively a small-scale application, and we configure 2 encoder layers and 1 decoder layer in \textsc{GiNet}. We implement \textsc{GiNet} in the PyTorch library. All experiments are conducted in our workstation with an AMD Ryzen 9 5950X processor and NVIDIA GTX 3080 GPU.

We present experiment results based on statistical performance. Among different performance evaluation metrics, we choose two mainstream ones including MAE and root mean squared error (RMSE). Relatively, RMSE is more susceptible to outliers than MAE because the mistakes are first squared. Let $\hat{\mathbf{y}}^\text{SoC} = \{\hat{y}^\text{SoC}_1, \ldots, \hat{y}^\text{SoC}_n\}$ be a list of $n$ SoC predictions and let $\mathbf{y}^\text{SoC} = \{y^\text{SoC}_1, \ldots, y^\text{SoC}_n\}$ be the corresponding ground-truth SoC. MAE is given by,
\begin{equation}
    \text{MAE} = \frac{1}{n} \sum\nolimits_{k=1}^{n} \left|\hat{y}_k^\text{SoC} - y_k^\text{SoC} \right|,
    \label{eq:mae}
\end{equation}
and RMSE can be calculated as,
\begin{equation}
    \text{RMSE} = \sqrt{\frac{1}{n} \sum\nolimits_{k=1}^{n} \left( \hat{y}_k^\text{SoC} - y_k^\text{SoC} \right)^2}.
    \label{eq:rmse}
\end{equation}

In the following parts, we show that our proposed \textsc{GiNet} achieves minimal MAE and RMSE for SoC prediction.

\subsection{Comparison Study}
We first demonstrate \textsc{GiNet}'s performance competitiveness by comparing \textsc{GiNet} with three comparison algorithms, including LSTM, GRU, and Informer. We have implemented comparison algorithms with parameters optimized and we consider different settings for a fair comparison. Specifically, we investigate the algorithms' performance with an input window of 10, 100, and 200 and a forecast horizon of 10 and 25. We report the performance in terms of MAE and RMSE and the results are shown in Table \ref{tab:comp}. We have the following observations. 

\begin{table}
\caption{A comparison study of \textsc{GiNet} and comparison algorithms including LSTM, GRU, and Informer for battery capacity prediction. Various combinations of input window and forecast horizon have been tested and the results based on MAE and RMSE are reported. The performance improvement in percentage with LSTM as the baseline shows \textsc{GiNet}'s significant error reduction for SoC prediction.}
\label{tab:comp}
\centering
\renewcommand{\arraystretch}{1.3}
\begin{tabular}{c|ccc|ccc|c}
\hline\hline
Forecast Horizon & \multicolumn{3}{c|}{10} & \multicolumn{3}{c|}{25} & 10 \\ \cline{2-8} 
Input Window & 10 & 100 & 200 & 10 & 100 & 200 & 200 \\ \cline{1-8}
SoC Model & \multicolumn{6}{c|}{MAE} & Impr. \\ \hline
LSTM & 0.33 & 0.27 & 0.26 & 0.36 & 0.27 & 0.26 & $-$ \\ \hline
GRU & 0.25 & 0.25 & 0.24 & 0.27 & 0.25 & 0.24 & 7.7 \\ \hline
Informer & 0.18 & 0.17 & 0.17 & 0.20 & 0.18 & 0.18 & 34.6 \\ \hline
\textsc{GiNet} (ours) & 0.15 & 0.14 & \textbf{0.11} & 0.19 & 0.15 & \textbf{0.13} & \textbf{57.7} \\ \hline
\hline 
SoC Model & \multicolumn{6}{c|}{RMSE} & Impr. \\ \hline
LSTM & 0.42 & 0.30 & 0.29 & 0.43 & 0.31 & 0.30 & $-$ \\ \hline
GRU & 0.29 & 0.28 & 0.26 & 0.30 & 0.29 & 0.27 & 10.3 \\ \hline
Informer & 0.21 & 0.21 & 0.19 & 0.22 & 0.21 & 0.20 & 34.5 \\ \hline
\textsc{GiNet} (ours) & 0.17 & 0.16 & \textbf{0.14} & 0.22 & 0.18 & \textbf{0.15} & \textbf{51.7} \\ \hline\hline
\end{tabular}
\end{table}

\subsubsection{Overall Results}
First, \textsc{GiNet} performs the best. In nearly all tested configurations of input window and forecast horizon, as well as for different metrics, \textsc{GiNet} outperforms the comparison algorithms. GRU and Informer are two building blocks of \textsc{GiNet}, and we notice that applying either GRU or Informer alone is insufficient for achieving optimal performance. This highlights \textsc{GiNet}'s concept of integrating sequential and context information, handled by GRU and Informer, respectively, for battery capacity prediction. Relatively, Informer alone performs better than GRU, and this demonstrates the superior modelling capability of the latest Transformer architecture. For sequential data analysis, LSTM is one of the most popular algorithms. Both LSTM and GRU are RNNs where GRU's architecture is simpler with fewer gates and parameters. We can see that LSTM incurs higher errors than GRU and the reason should be that GRU's simple architecture aligns well with our application, which is relatively small-scale with over-fitting risks. Let LSTM be the baseline algorithm. We calculate the error reduction improvement for the rest algorithms with input window 200 and forecast horizon 10. We can see that GRU and Informer outperform LSTM with 8\% and 35\% lower MAE, respectively. \textsc{GiNet} has the best performance and the improvement is 58\%, which is 46\% and 28\% better than its building blocks GRU and Informer, respectively. RMSE results support our observations as well.

\subsubsection{Input Window}
A large input window helps improve performance. Many ML models favor more data to establish an accurate mapping from the input data to the output. For battery capacity prediction, the prediction output is correlated with not only short-term historical battery operation data but also long-term data. \textsc{GiNet} employs sequential learning and context-aware learning for short-term and long-term data, respectively. Given a large input window, our tested algorithms have more information to process and can achieve improved performance if the information is processed well. For \textsc{GiNet} with forecast horizon 10, the achieved MAE is 0.15, 0.14, and 0.11 for input window 10, 100, and 200, respectively, with a significant MAE improvement of 27\% from window 10 to 200. LSTM also shows a 21\% improvement from window 10 to 200 for MAE. Interestingly, we observe that GRU or Informer alone is much less sensitive to the input window size compared to \textsc{GiNet}, where the difference of MAE is around 0.01 only for forecast horizon 10. This implies the importance of dealing with both short-term and long-term data with different technologies to well explore the potential of large window input data.

\subsubsection{Forecast Horizon}
Besides, predicting for a long horizon is challenging. Compared to horizon 10, both MAE and RMSE results of the tested algorithms for horizon 25 are generally higher. For \textsc{GiNet} with input window 200, its MAE is 18\% higher with the forecast increased from 10 to 25. Intuitively, the historical battery data is more correlated to the near future battery dynamics and the long-term changes depend on the updated battery operation information which is not yet available. Given that the prediction accuracy drops for long horizons, it is important to convey the algorithm strength and limitation to users with comprehensive performance measurements, e.g., by quantifying the prediction uncertainty \cite{10757952}.



\subsection{Sensitivity Analysis}
\textsc{GiNet} has several parameters that shall be optimized for optimal performance. In this part, we conduct sensitivity analysis to understand \textsc{GiNet}'s performance variation with different parameter settings. Specifically, we investigate \textsc{GiNet}'s \texttt{ProbSparse} attention mechanism, distillation, and the number of encoder and decoder layers.

\begin{table}
\caption{\textsc{GiNet}'s performance sensitivity to different Informer configurations for attention mechanism and distillation. The results are generated with input window 100 and forecast horizon 25 and both MAE and RMSE results are reported. \textsc{GiNet} with \texttt{ProbSparse}-based attention and distillation enabled has the optimal performance for SoC prediction.}
\label{tab:sens}
\centering
\renewcommand{\arraystretch}{1.3}
\begin{tabular}{c|cc|cc}
\hline\hline
\multirow{2}{*}{Attention Mechanism} & \multicolumn{2}{c|}{w/ distillation} & \multicolumn{2}{c}{w/t distillation} \\ \cline{2-5} 
 & MAE & RMSE & MAE & RMSE \\ \cline{1-5}
\texttt{ProbSparse} & \textbf{0.15} & \textbf{0.18} & 0.16 & 0.19 \\ \hline
Full Attention & 0.16 & 0.19 & 0.18 & 0.21 \\ \hline
\hline
\end{tabular}
\end{table}

\subsubsection{Attention}
\textsc{GiNet} uses \texttt{ProbSparse} attention which addresses the inefficiencies of full attention used in standard Transformer models by focusing on the most critical dependencies within the input sequence. In full attention, all data points in a sequence are treated equally and all pairwise interactions are calculated. \texttt{ProbSparse} is different by selecting informative interactions only and enables the model to use memory efficiently and focus on the most relevant battery patterns for effective learning. Seen from Table \ref{tab:sens}, \texttt{ProbSparse} helps reduce the prediction error compared to full attention and improve MAE by 6.3\% and 5.3\% for the implementation with and without distillation, respectively. 

\subsubsection{Distillation}
Distillation complements \texttt{ProbSparse} for efficient and effective processing of long battery sequences. While \texttt{ProbSparse} focuses on optimizing attention, distillation aims to reduce the length of the data input. It applies convolutional layers to shorten the data sequence in the encoder without sacrificing temporal and contextual information. With reduced redundancy, the distilled data allows \textsc{GiNet} to focus on high-level features for improved generalization and performance on new battery dynamics. Table \ref{tab:sens} shows that distillation helps improve prediction accuracy for both \texttt{ProbSparse}-based and full attention, where the MAE improvement is 6.3\% and 11.1\%, respectively.

\subsubsection{Encoder Decoder Layers}
\begin{figure}
    \centering	\includegraphics[width=0.98\linewidth]{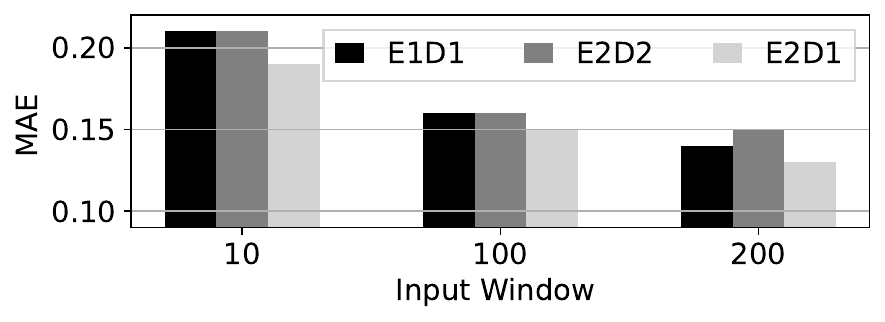}
    \caption{The \textsc{GiNet}'s achieved MAE with different numbers of encoder and decoder layers. The results for input windows 10, 100, and 200 for forecast horizon 25 are reported. E$i$D$j$ in legend means $i$ encoder layers and $j$ decoder layers. \textsc{GiNet} performs the best with two encoder layers and one decoder layer, i.e., E2D1 in the figure.}
    \label{fig:layer}
\end{figure}

Encoder and decoder are key components of \textsc{GiNet} (refer to Fig. \ref{fig:ginet}). Inside each encoder, there can be multiple layers, each with attention, feedforward layer, distillation, etc., and the multiple layers can be stacked together in the encoder. This is true for the decoder also, except for some differences such as the decoder's specialized attention. In this part, we test the performance of \textsc{GiNet} with different numbers of encoder layers and decoder layers. As battery capacity modelling is relatively a small-scale application, we configure the \textsc{GiNet} models with minimal encoder and decoder layers. Specifically, we test three configurations, including one encoder layer and one decoder layer, two encoder layers, and two decoder layers, and two encoder layers and one decoder layer, denoted as E1D1, E2D2, and E2D1, respectively, as shown in Fig. \ref{fig:layer}. We can see that one encoder layer is insufficient to achieve optimal performance. Compared to the best configuration, i.e., E2D1 with two encoder layers and one decoder layer, the MAE is on average 8.3\% higher for one encoder layer only. This suggests that our battery data embedding is not simple enough to be well modeled with one encoder layer. An encoder shall well process \textsc{GiNet}'s embedding that contains a comprehensive representation of both original battery data and the GRU-generated temporal dependencies. For the decoder, we notice that one layer supports optimal performance, and increasing the number of decoder layers does not enhance \textsc{GiNet}'s prediction, with 10.9\% higher MAE on average. A potential reason is that the decoder's role in \textsc{GiNet} is to interpret encoder-generated representations and produce battery capacity predictions and one layer meets the application requirements well. When more layers are configured for the decoder, there could be unnecessary model complexity and risks of overfitting. 
 
\balance
\section{Conclusion}
\label{sec:conclusion}
In this paper, we propose \textsc{GiNet}, a GRU-enhanced Informer network for battery SoC prediction. \textsc{GiNet} is designed and developed to capture both sequential and contextual information of battery dynamics where battery capacity changes are correlated with both information. Among the \textsc{GiNet} components, GRU models sequential dependencies and Informer captures contextual information by analyzing long sequence battery data with its attention mechanism. \textsc{GiNet} inherits and customizes Informer's \texttt{ProbSparse} attention and distillation operation for efficient and effective battery data processing. An experimental study is conducted based on a public dataset with Panasonic 18650PF battery cells. We demonstrate that \textsc{GiNet} yields a low error rate and stable prediction across various input window and forecast horizon configurations. The minimal achieved MAE for \textsc{GiNet} is 0.11 and \textsc{GiNet} outperforms comparison algorithms, including LSTM, GRU, and Informer, significantly. The average error reduction compared to the second best performed Informer is 27\%. And RMSE results suggest similar insights. 

In the future, we would like to enhance the extraction of sequential and contextual information with further optimized GRU and Informer or the latest ML algorithms. Feature fusion plays a vital role in ML \cite{kwan2024nutritionestimationdietarymanagement} and we plan to upgrade \textsc{GiNet}'s feature fusion module to further enhance the performance. Finally, we would like to improve \textsc{GiNet}'s generalizability of processing other battery datasets and customize \textsc{GiNet} for other battery applications such as battery health monitoring.

\bibliographystyle{IEEEtran}
\bibliography{reference}

\end{document}